\pdfoutput=1
\documentclass[11pt]{article}

\usepackage{authblk}
\usepackage{ragged2e}
\usepackage[final]{acl}

\usepackage{times}
\usepackage{latexsym}

\usepackage[T1]{fontenc}
\usepackage[utf8]{inputenc}
\usepackage{microtype}
\usepackage{inconsolata}
\usepackage{graphicx}

\usepackage{booktabs}
\usepackage{multirow}
\usepackage{combelow}
\usepackage{tabularx}
\usepackage{xcolor}
\usepackage{lipsum}
\usepackage{tabularray}

\usepackage{float}
\usepackage{placeins}
\title{RoQLlama: A Lightweight Romanian Adapted Language Model}

\author[1]{\textbf{George-Andrei Dima}}
\author[1]{\textbf{Andrei-Marius Avram}}
\author[1,2]{\textbf{Cristian-George Cr\u{a}ciun}}
\author[1]{\\ \textbf{Dumitru-Clementin Cercel}\thanks{Corresponding author: dumitru.cercel@upb.ro.}}
\affil[1]{National University of Science and Technology POLITEHNICA Bucharest, Romania}
\affil[2]{Technical University Munich, Munich, Germany}

\begin{document}
\maketitle

\begin{abstract}
The remarkable achievements obtained by open-source large language models (LLMs) in recent years have predominantly been concentrated on tasks involving the English language. In this paper, we aim to advance the performance of Llama2 models on Romanian tasks. We tackle the problem of reduced computing resources by using QLoRA for training. We release RoQLlama-7b, a quantized LLM, which shows equal or improved results compared to its full-sized counterpart when tested on seven Romanian downstream tasks in the zero-shot setup. Also, it consistently achieves higher average scores across all few-shot prompts. Additionally, we introduce a novel Romanian dataset, namely RoMedQA, which contains single-choice medical questions in Romanian.
\end{abstract}

\section{Introduction} 


Transformer models \citep{vaswani2017attention} represent the state-of-the-art solution adopted by natural language processing (NLP) tasks \cite{wilie2020indonlu}. Due to current breakthroughs in computational capabilities, models were scaled in terms of parameters, acquiring new remarkable abilities in terms of natural language understanding \cite{elmadany2023orca}. As a result, a series of proprietary and open large language models (LLMs) were created.


One major downside of LLMs is the enormous amount of computational resources and training data they require. 
Democratizing LLMs constitutes a vital research direction, increasing the possibility of breakthroughs. In this sense, we release a new lightweight Romanian language-adapted LLM with 7 billion parameters and quantized to 4 bits by employing the state-of-the-art quantized LoRA (QLoRA) training technique \citep{dettmers2023qlora}. We evaluate our model on several Romanian datasets, covering seven tasks and comparing it to its original counterpart. Our results showed that RoQLlama-7b outperformed the other Llama models on four out of the seven tasks investigated using zero-shot prompting. Furthermore, due to quantization, the model has a significantly smaller memory footprint, up to three times less than the base model.

To summarize, the contributions of this work are:
\begin{itemize}
    \item We train and release the first Romanian-adapted LLM based on Llama2-7b \cite{touvron2023llama2}, with reduced memory footprint, called \textbf{RoQLlama-7b}\footnote{\url{https://huggingface.co/andreidima/Llama-2-7b-Romanian-qlora}}.
    \item We introduce \textbf{RoMedQA}\footnote{\url{https://huggingface.co/datasets/craciuncg/RoMedQA_v1}}, the first dataset comprising medical exam questions in the Romanian language.
    \item We comprehensively test the released model, comparing it with the Llama2-7b models on Romanian downstream tasks.
    \item We investigate parameter efficiency and language adaptation in a low-resource language setting. 
\end{itemize}

\section{RoQLlama}

\subsection{Training Dataset}

When building our training data, we start from the work done by \citet{masala-etal-2020-robert}. We use RoWiki, a Romanian Wikipedia dump containing 0.3 GB of text, and RoTex, a text collection from online Romanian sources containing 1.5 GB of text. 

Also, we included in our training data the Romanian sections from the OSCAR corpus \citep{suarez2019asynchronous}, containing 45.6 GB of text, and from the CC-100 corpus \citep{conneau-etal-2020-unsupervised}, containing 61.4 GB of text. \citet{touvron2023llama} suggest that various pre-processed CommonCrawl \cite{smith2013dirt} variants could enhance the obtained results. Therefore, we decided to use the Romanian corpora from both OSCAR and CC-100 in our training data, even though both are based on the same data source. See Appendix \ref{app:train_ds_processsing} for information on the steps involved in processing the training dataset.

\subsection{Training Process}

We trained the model using QLoRA. We take the advice from \cite{dettmers2023qlora} regarding the low-rank adaptation (LoRA) \citep{hu2021lora} adapter hyperparameters, as they believe that if LoRA is applied to every layer, \textit{LoRA r} will not impact the experimental results.
We applied LoRA to all linear layers of Llama2 and set \textit{LoRA r} at 8. We also kept \textit{LoRA alpha} at 8 and set \textit{LoRA dropout} at 0.05 since dropout has been shown to boost performance in the smaller Llama variants \citep{dettmers2023qlora}. Regarding the quantization of the base models, we used the 4-bit NF4 quantization and did not apply double quantization.

We used the paged AdamW optimizer \cite{loshchilov2017decoupled} with a learning rate of 1e-5, a weight decay of 0.001, and a gradient clipping set at 0.01. In order to fit the graphics processing unit (GPU), we use a per-device batch size of 2 with 4 gradient accumulation steps, resulting in a batch size of 8. The model was trained for 900,000 steps, which is approximately 7.3B tokens.

\subsection{Memory Footprint}

\begin{table}[t]
  \centering
  \begin{tabular}{lccc}
    \toprule
    \textbf{Model} & \textbf{M1 (GB)} & \textbf{M2 (GB)} \\
    \midrule
     Llama2-7b & 13.4 & 14.8 \\
     \textit{RoQLlama-7b} & 4.7 & 6.1 \\
     \bottomrule
  \end{tabular}
  \caption{Memory footprints of Llama2-7b and RoQLlama-7b.
  M1 represents the VRAM used by the model, whereas
  M2 represents the VRAM needed to ingest a prompt of 1,000 tokens.}
  \label{tab:resources}
\end{table}

In Table \ref{tab:resources}, we compare our quantized model with the original version regarding memory footprint and processing time. M1 represents the video random-access memory (VRAM) usage measured after loading each model into the GPU, with the original model loaded in float16. M2 indicates the maximum VRAM used when processing an artificial prompt of 1,000 tokens. All tests were conducted on an A100 80GB NVIDIA GPU.

Our model has a significantly smaller memory footprint, reducing both the M1 and M2 memory required to run it from 13.4 GB to 4.7 GB and from 14.8 GB to 6.1 GB, respectively.

\section {RoMedQA}
More publicly available datasets are needed for the Romanian NLP tasks. To contribute to this area, we introduce \textbf{RoMedQA}, a dataset that amounts to 4,127 single-choice questions regarding the medical field in the Romanian language. The dataset consists of advanced biology questions used in entrance examinations in medical schools in Romania. Each question has five possible answer choices, numbered from 1 to 5, with only one correct answer. See Appendix \ref{app:romedqa} for a more detailed dataset description.

\section{Evaluation}

\begin{table*}[t]
  \centering
  \resizebox{\textwidth}{!}{
  \begin{tabular}{lcccccccc}
    \toprule
    \textbf{Model} & \textbf{RoMedQA} & \textbf{RoQA} & \textbf{REDv2} & \textbf{RoMD} & \textbf{SaRoCo} & \textbf{RoSum} & \textbf{RoSTS} & \textbf{Avg.} \\
    \midrule
     Llama2-7b & 3.60 & 24.88 & 3.59 & 4.95 & 28.17 & 18.47 & -0.663 & 14.36 \\
     Llama2-7b-chat & 1.79 & \textbf{44.05} & \textbf{6.89} & 20.38 & \textbf{29.88} & 22.26 & 0.039 & 25.31 \\
     \midrule
     \textit{RoQLlama-7b} & \textbf{3.67} & 39.64 & 6.45 & \textbf{29.78} & 29.63 & \textbf{19.46} & \textbf{0.401} & \textbf{28.38} \\
     \bottomrule
  \end{tabular}
  }
  \caption{Zero-shot results of the Llama2 models on all the 7 evaluated tasks, together with the average score.}
  \label{tab:zeroshot_all}
\end{table*}

\begin{figure*}
    \centering
    \includegraphics[width=0.95\textwidth]{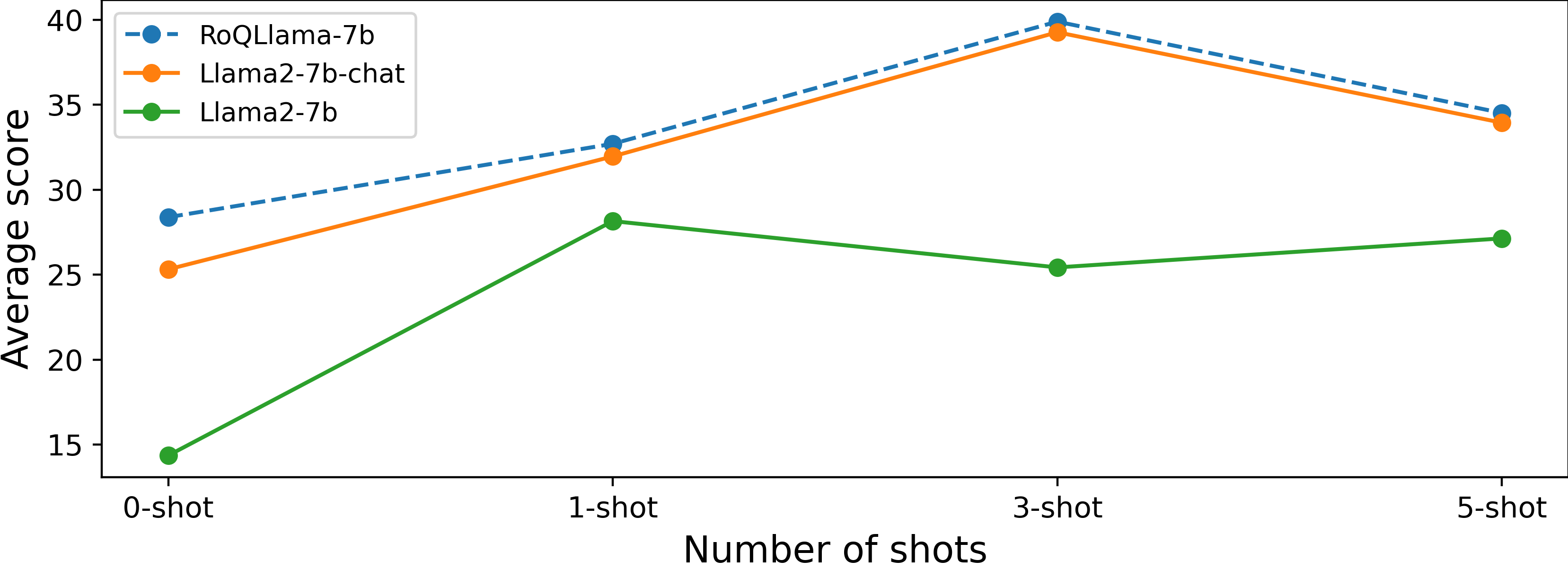}
    \caption{Average few-shot results of the Llama2 models.}
    \label{fig:few-shot}
\end{figure*}

We evaluate the RoQLlama-7b model and the original Llama2 on seven Romanian NLP tasks as follows:
\begin{itemize}
    \item \textbf{Medical Question Answering} - using the new dataset (RoMedQA) introduced in this work.
    \item \textbf{Question Answering} - using the Romanian subset (RoQA) \cite{dumitrescu2021liro} of the Cross-Lingual Question Answering Dataset (xSQuAD) \cite{artetxe2020cross}.
    \item \textbf{Emotion Detection} - using the second version of Romanian Emotion Dataset (REDv2) \cite{ciobotaru2022red}.
    \item \textbf{Romanian/Moldavian Dialect Classification} - using the Moldavian and Romanian dialectal Corpus (MOROCO) \cite{butnaru2019moroco}.
    \item \textbf{Satire Detection} - using the Satire detection Romanian Corpus (SaRoCo) \cite{rogoz2021saroco}.
    \item \textbf{News Summarization} - on the Romanian Summarization dataset (RoSum) \cite{niculescu2022rosummary}.
    \item \textbf{Textual Similarity} - using the Romanian Semantic Textual Similarity dataset (RoSTS) \cite{dumitrescu2021liro}
\end{itemize}

We compare them to the original baselines, which include Ro-BERT \cite{dumitrescu2020birth}, Ro-GPT2 \cite{niculescu2021rogpt2}, and various other architectures. During the evaluation process, we kept the same configuration for all models and all tasks:
\textit{temperature} = 0.6, and 
\textit{top\_p} = 0.9. Also, we stopped the generation at the first new line. 
We varied the maximum number of generated tokens for different task categories: 10 for classification and regression tasks (RoMedQA, REDv2, RoMD, SaRoCo, RoSTS), 250 for question answering (RoQA), and 2,048 for summarization (RoSum).

We show the original prompts used to evaluate the language models on each task, as well as their translation in English in Appendix \ref{app:eval_prompts}.

\section{Results}

The results using zero-shot prompting are depicted in Table \ref{tab:zeroshot_all}, which outlines the macro F1-scores on each classification task (i.e., RoMedQA, REDv2, RoMD, and SaRoCo), the overlap F1-score on the question answering task (i.e., RoQA), the ROUGE-L score on the summarisation task (i.e., RoSum), and the Pearson correlation score for the regression task (i.e., RoSTS). The RoQLlama-7b model obtained the highest score on four out of the seven evaluated tasks, namely on RoMedQA, RoMD, RoSum, and RoSTS. The highest average score of 28.39 was obtained by Llama2-7b, outperforming Llama2-7b-chat by $\sim$3\% and almost doubling the average score\footnote{To compute the average score, we normalized the Pearson score which we obtained on RoSTS.}.

We also evaluate the performance of the newly introduced RoQLlama-7b model on few-shot prompting (i.e., one-shot, three-shot, and five-shot). This allows us to analyze how the model improves when additional examples for each task are given. The results of this analysis are depicted in Figure \ref{fig:few-shot}, which outlines the variations in average scores when an increasing number of examples are presented to each model in the prompt. We can observe that RoQLlama-7b consistently outperforms both Llama2-7b and Llama2-7b-chat on all the few shots prompting tasks evaluated in this work.

Furthermore, because the language models can hallucinate and produce inadequate output for each classification task, we also evaluate the \textit{not followed instruction} (NFI) score, which measures the percentage of samples on which the model fails to adhere to the given instructions. The following subsections present results on zero-shot prompting for each evaluation task.

\subsection{RoMedQA}

To establish a baseline for comparing these results with other future results that can be achieved by training, we evaluate the test split of the RoMedQA, which contains 831 entries.

As shown in Table \ref{tab:romedqa}, RoQLlama-7b outperforms both Llama2-7b models regarding macro F1-score. However, the scores of all three models remain unsatisfactory. This is expected, as none of the models are trained explicitly on medical data, which is essential for solving the questions in the dataset.

\begin{table}[t]
  \centering
  \begin{tabular}{lccc}
    \toprule
    \textbf{Model} & \textbf{Acc} & \textbf{F1} & \textbf{NFI}\\
    \midrule
     Llama2-7b & \textbf{22.77} & 3.60 & 0.00 \\
     Llama2-7b-chat & 11.69 & 1.79 & 0.00 \\
     \midrule
     \textit{RoQLlama-7b} & 21.57 & \textbf{3.67} & 0.00 \\
     \bottomrule
  \end{tabular}
  \caption{Evaluation results of our models on the RoMedQA dataset. The scores of Llama2 models are calculated using zero-shot prompting.}
  \label{tab:romedqa}
\end{table}

\subsection{RoQA}

\begin{table}[t]
  \centering
  \begin{tabular}{lcc}
    \toprule
    \textbf{Model} & \textbf{EM} & \textbf{F1} \\
    \midrule
     mBERT & 58.99 & 72.69 \\
     XLM-R Large & 69.66 & 83.56 \\
     \midrule
     Llama2-7b & 13.94 & 24.88 \\
     Llama2-7b-chat & \textbf{25.12} & \textbf{44.05} \\
     \midrule
     \textit{RoQLlama-7b} & 24.20 & 39.64 \\
     \bottomrule
  \end{tabular}
  \caption{Evaluation results of our models on the Romanian xSQuAD subset. The scores of Llama2 models are calculated using zero-shot prompting. Non-Llama baselines are taken from \cite{dumitrescu2021liro}.}
  \label{tab:xsquad}
\end{table}

We evaluate the original and the Romanian Llama2 models on the RoQA, which contains 240 paragraphs and 1,190 question-answer pairs annotated using the SQuAD v1.1 guidelines \cite{rajpurkar2016squad}. Table \ref{tab:xsquad} depicts the results with zero-shot prompting. With an Exact Match (EM) score of 24.20 and an overlap F1-score of 39.64, RoQLlama-7b performs better than its original counterpart in terms of both EM and overlap F1-score, outperforming the Llama2-7b model by a considerable margin. However, its performance is slightly worse than that of the Llama2-7b-chat.

\subsection{REDv2}

We evaluate the REDv2 dataset, which contains collected tweets in the Romanian language, annotated with their associated emotions. RoQLlama-7b performs better than Llama2-7b in terms of macro F1-score and accuracy. However, the highest accuracy is obtained by the Llama2-7b-chat model, with 48.24\%, almost double that of the other two models. Llama2-7b-chat also obtains the highest macro F1-score on REDv2.

\begin{table}[t]
  \centering
  \begin{tabular}{lccc}
    \toprule
    \textbf{Model} & \textbf{Acc} & \textbf{F1} & \textbf{NFI}\\
    \midrule
     Ro-BERT & 54.1 & 66.8 & - \\
     XLM-RoBERTa & 50.4 & 61.9 & - \\
     \midrule
     Llama2-7b & 26.46 & 3.59 & 0.00 \\
     Llama2-7b-chat & \textbf{48.24} & \textbf{6.89} & 0.00 \\
     \midrule
     \textit{RoQLlama-7b} & 26.34 & 6.45 & 0.00 \\
     \bottomrule
  \end{tabular}
  \caption{Evaluation results of our models on the REDv2 dataset. The scores of Llama2 models are calculated using zero-shot prompting. Non-Llama baselines are taken from \cite{ciobotaru2022red}.}
  \label{tab:redv2}
\end{table}

\subsection{RoMD}

We evaluate the Romanian Llama2 model introduced in this work on a classification task to determine whether a text belongs to the Romanian or Moldavian dialects using the test subset of the MOROCO. The results are depicted in Figure \ref{tab:romd}. RoQLlama-7b obtains the highest accuracy and macro F1-score out of all Llama2 models, with a 46.84\% accuracy and a 29.78\% macro F1-score. Also, it shows the lowest NFI score, the model not following instructions in 11.71\% of the cases.

\begin{table}[t]
  \centering
  \begin{tabular}{lccc}
    \toprule
    \textbf{Model} & \textbf{Acc} & \textbf{F1} & \textbf{NFI}\\
    \midrule
     KRR + $k_6^{0/1}$ & 94.13 & 94.06 & - \\
     CNN & 92.75 & 92.71 & - \\
     CNN + SE & 92.99 & 92.93 & - \\
     \midrule
     Llama2-7b & 4.42 & 4.95 & 91.36 \\
     Llama2-7b-chat & 30.58 & 20.38 & 42.01 \\
     \midrule
     \textit{RoQLlama-7b} & \textbf{46.84} & \textbf{29.78} & \textbf{11.71} \\
     \bottomrule
  \end{tabular}
  \caption{Evaluation results of our models on the RoMD dataset. The scores of Llama2 models are calculated using zero-shot prompting. Non-Llama baselines are taken from \cite{butnaru2019moroco}.}
  \label{tab:romd}
\end{table}

\subsection{SaRoCo}

The SaRoCo introduces a dataset designed for identifying satirical content in Romanian news articles \cite{rogoz2021saroco}. The results depicted in Table \ref{tab:saroco} outline that Llama2-7b-chat outperformed RoQLlama-7b both in terms of accuracy, where it achieved a score of 50.12\% and macro F1-score with 29.88\%. However, RoQLlama-7b obtained the lowest NFI score out of all three models, with 7.98\% of the answers being inadequate concerning the provided instructions.

\begin{table}[t]
  \centering
  \begin{tabular}{lccc}
    \toprule
    \textbf{Model} & \textbf{Acc} & \textbf{F1} & \textbf{NFI}\\
    \midrule
     Ro-BERT & 73.00 & 71.50 & - \\
     Char-CNN & 69.66 & 71.09 & - \\
     \midrule
     Llama2-7b & 11.69 & 28.17 & 21.27 \\
     Llama2-7b-chat & \textbf{50.12} & \textbf{29.88} & 8.72 \\
     \midrule
     \textit{RoQLlama-7b} & 41.12 & 29.63 & \textbf{7.98} \\
     \bottomrule
  \end{tabular}
  \caption{Evaluation results of our models on the SaRoCo dataset. The scores of Llama2 models are calculated using zero-shot prompting. Non-Llama baselines are taken from \cite{butnaru2019moroco}.}
  \label{tab:saroco}
\end{table}

\subsection{RoSum}

We compare RoQLlama-7b with the original Llama2 models on RoSum for summarization performance. This summarization dataset was created by crawling Romanian news articles, which also provided bullet point summaries. Table \ref{tab:rosum} presents the results, which indicate that Llama2-7b-chat and, to a lesser extent, Llama2-7b outperform our model. This may be due to our model's training being conducted on relatively small samples of Romanian text, each with fewer than 1,024 tokens.

\begin{table}[t]
  \centering
  \begin{tabular}{lccc}
    \toprule
    \textbf{Model} & \textbf{R-1} & \textbf{R-2} & \textbf{R-L} \\
    \midrule
    Ro-GPT2-base & 34.80 & 19.91 & 34.16 \\
    Ro-GPT2-medium & 35.46 & 20.61 & 34.67 \\
    Ro-GPT2-large & 34.92 & 19.95 & 33.84 \\
    \midrule
     Llama2-7b & 24.67 & 12.22 & 18.47 \\
     Llama2-7b-chat & \textbf{32.05} & \textbf{14.72} & \textbf{22.26} \\
     \midrule
     \textit{RoQLlama-7b} & 24.37 & 11.70 & 18.46 \\
     \bottomrule
  \end{tabular}
  \caption{Evaluation results of our models on the Romanian summarization dataset. The scores of Llama2 models are calculated using zero-shot prompting. Non-Llama baselines are taken from \cite{niculescu2022rosummary}.}
  \label{tab:rosum}
\end{table}

\subsection{RoSTS}

The original and Romanian Llama2 models were also evaluated on their performance for textual similarity using the test set of the RoSTS dataset, which contains 1,379 sentence pairs, each annotated with a similarity score from 0 to 5. We compare the models by computing both the Pearson and Spearman correlation coefficients. The results are depicted in Table \ref{tab:rosts}.  RoQLlama-7b performs significantly better on this task than both Llama2-7b and Llama-7b-chat, with the highest Pearson correlation (0.412) and the highest Spearman correlation (0.462).

\begin{table}[t]
  \centering
  \begin{tabular}{lcc}
    \toprule
    \textbf{Model} & \textbf{Pearson} & \textbf{Spearman} \\
    \midrule
     RNN & 0.685 & - \\
     mBERT (cased) & 0.766 & - \\
     mBERT (uncased) & 0.769 & - \\
     Ro-BERT (cased) & 0.792 & - \\
     Ro-BERT (uncased) & 0.815 & - \\
     \midrule
     Llama2-7b & -0.663 & -0.541 \\
     Llama2-7b-chat & 0.039 & 0.055 \\
     \midrule
     \textit{RoQLlama-7b} & \textbf{0.401} & \textbf{0.462} \\
     \bottomrule
  \end{tabular}
  \caption{Evaluation results of our models on the RoSTS test set. The scores of Llama2 models are calculated using zero-shot prompting. Non-Llama baselines are taken from \cite{dumitrescu2021liro}.}
  \label{tab:rosts}
\end{table}

\section{Conclusions}
In this paper, we advance state-of-the-art NLP techniques for Romanian and address the scarcity of Romanian datasets. We introduce a lightweight LLM for Romanian and a new medical dataset of single-choice exam questions. RoQLlama-7b, a quantized version of Llama2-7b, achieves higher average scores across Romanian tasks while using three times less memory.

RoMedQA is the first Romanian dataset of medical questions and answers based on entrance exams for medical schools in Romanian. It is valuable for training and testing LLMs in medical knowledge and language comprehension.
Future work includes enhancing the dataset with contextual information, adapting it for smaller models, and integrating the results into the Romanian LiRo benchmark \cite{dumitrescu2021liro}.

\section*{Limitations}
Since our model is a fine-tuned version of the Llama2 model, it inherits the existing limitations of the parent model, as shown by \citet{touvron2023llama2}. Additionally, our model was further trained on Internet text so that it may have been exposed to specific biases prevalent on the Romanian Internet. Users should be aware that this model carries risks of generating hallucinations, toxic language, and various biases.



\bibliography{ro_llama_bib}


\appendix

\section{Related Work}

\paragraph{English Llama Models.} Llama \citep{touvron2023llama} is a family of large language models trained exclusively on publicly available data and released openly, ranging in sizes from 7 to 65 billion parameters and with a context window of 2048 tokens. Llama2 \citep{touvron2023llama2} was introduced as a collection of large language models which showed improved performance compared to the first generation. Llama2 models have sizes ranging from 7 to 70 billion parameters and a context window of 4096 tokens. The versatility and accessibility of the models from the Llama family render them some of the most popular large language models. 

The first-generation and second-generation Llama models were the foundation for numerous research papers. For instance,  \citet{xie2023pixiu} fine-tuned Llama1 on an instruction dataset containing various tasks from the financial domain, showing the potential for domain-specific tuning of large language models. \citet{yuan2024continued} further trained Llama2-13b on texts from the medical domain in order to approach the problem of medical text generation.

\paragraph{Llama Models in Other Languages.} Adapting Llama models to low-resource languages has significantly improved downstream task performance for those languages.
\citet{Pires_2023} further trained Llama1 (sizes 7B and 65B) on a Portuguese corpus containing 7.3 billion tokens, resulting in better performance on Portuguese tasks.
Similarly, \citet{cui2024efficient} trained Llama 7B, 13B, and 33B on a Chinese corpus, achieving superior results in Chinese text comprehension and generation compared to the original model. \citet{kuulmets2024teaching} adapted Llama2 on Estonian while maintaining its performance in English by training on a combined English and Estonian corpus of 5 billion tokens. 
Also, many other languages have benefited from adapted versions of Llama, such as French \citep{gesnouin2024llamandement}, Italian \citep{santilli2023camoscio, basile2023llamantino}, Japanese \cite{enomoto2024investigating}, Galician \cite{gamallo2024open}, and
Vietnamese \citep{nguyen2023vinallama}.

\paragraph{QLoRA.} Parameter efficient fine-tuning (PEFT) techniques \citep{peft, xu2023parameterefficient} aim to adapt large models for specific tasks with minimal computational resources, addressing the challenges posed by their enormous scale. A well-known PEFT method is LoRA \citep{hu2021lora}, which reduces the resources needed to train large language models by training only the rank-decomposition matrices corresponding to the dense layers of the Transformer architecture instead of full parameter training. Models trained with LoRA have been shown to require up to three times less memory than usual training with very little to almost no loss in performance \citep{hu2021lora}.

QLoRA \citep{dettmers2023qlora} further reduces the required resources by introducing 4-bit normal float quantization (NF4), double quantization (DQ), and paged optimizers. NF4 is a novel quantization technique that goes beyond a crude model approximation and cleverly uses the available 4 bits to minimize the loss of information in the model's parameters. Using the normal distribution, NF4 eliminates outlier parameters and accurately represents the more often occurring parameter values. DQ further achieves memory savings by quantizing the quantization constants. The memory overhead caused by quantization constants is typically 0.5 bits per parameter. When using DQ, quantization constants have a memory footprint of 0.127 bits per parameter.

Building on the advantages of PEFT training, numerous research papers have focused on fine-tuning Llama models with LoRA or QLoRA. \citet{gema2023parameterefficient} fine-tuned Llama for the clinical domain using LoRA and reported state-of-the-art results across clinical tasks.
\citet{zhang2024tablellama} addressed tasks based on semi-structured tables by building TableLama, a Llama2-7b model fine-tuned with LongLoRA \cite{chen2023longlora}.
\citet{santilli2023camoscio} used LoRA to fine-tune Llama on a corpus of instructions translated into Italian and reported competitive results on Italian downstream tasks.
\citet{basile2023llamantino} built LLaMAntino by adapting the Llama2-7b and 13B models to the Italian language using QLoRA.

\section{RoMedQA}
\label{app:romedqa}

\paragraph{Overview.} One of the significant issues in the Romanian NLP tasks is the need for more available data. We decided to make our contribution based on several works \citep{wang2022archivalqa,keren2021parashoot,huang2019geosqa} performed by various linguistic research communities that enriched data availability in their respective languages, opening the doors to new research possibilities in terms of intra and inter-lingual NLP. In light of the above, we introduce \textbf{RoMedQA}, a dataset that amounts to 4,127 single-choice questions regarding the medical field in the Romanian language. The dataset consists of advanced biology questions used in entrance examinations in medical schools in Romania.


\begin{table}
  \centering
  \begin{tabular}{l|l|c}
    \toprule
    \textbf{Word} & \textbf{Translation} & \textbf{TF-IDF Score}\\
    \midrule
     celulă & cell & 0.02205 \\
     mușchi & muscle & 0.02192 \\
     nerv & nerve & 0.02026 \\
     nivel & level & 0.01998 \\
     corect & correct & 0.01991 \\
     răspuns & answer & 0.01935 \\
     niciun & none & 0.01895 \\
     afirmație & statement & 0.01848 \\
     fibră & fiber & 0.01810 \\
     următor & next & 0.01754 \\
     \bottomrule
  \end{tabular}
  \caption{TF-IDF scores of the most common words found in RoMedQA.}
  \label{table:tfidf}
\end{table}

\paragraph{Data Collection.}
Building this dataset was quite challenging because of the variety in the data format. We had to resort to multiple techniques to collect the entries. The questions given at past entrance examinations were available on the medical universities' websites in HTML format, PDF documents, or scans. Where possible, we used web scrapping to extract the questions with their respective correct answer. In other cases, we had to scrape PDF documents for text or, less favorably, perform OCR on the PDF scans to extract the underlying questions and answers. 

Unfortunately, some scans were not of good quality. Therefore, we manually extracted the questions written on the scanned documents. Ultimately, we manually inspected the data to identify and rectify any noise introduced by the OCR process. This ensures that our dataset is of good quality, with no issues for anyone to use, eliminating the need for sanitization and other data pre-processing. 

\begin{figure}[H]
    \centering
    \includegraphics[width=0.9\columnwidth]
    {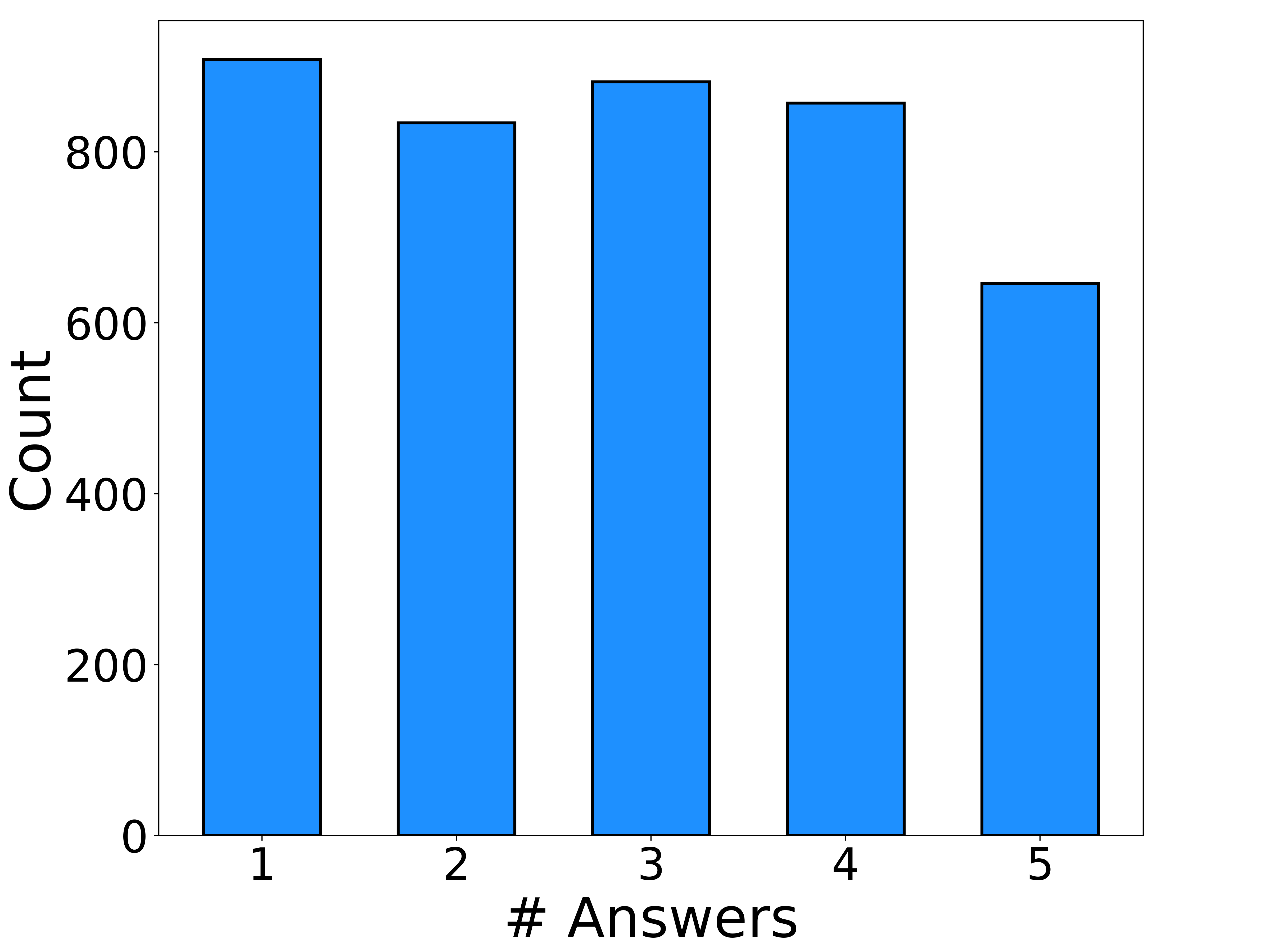}
    \caption{Class distribution of the RoMedQA dataset.}
    \label{fig:class}
\end{figure}

\begin{figure*}[!ht]
    \centering
    \includegraphics[width=0.75\textwidth]{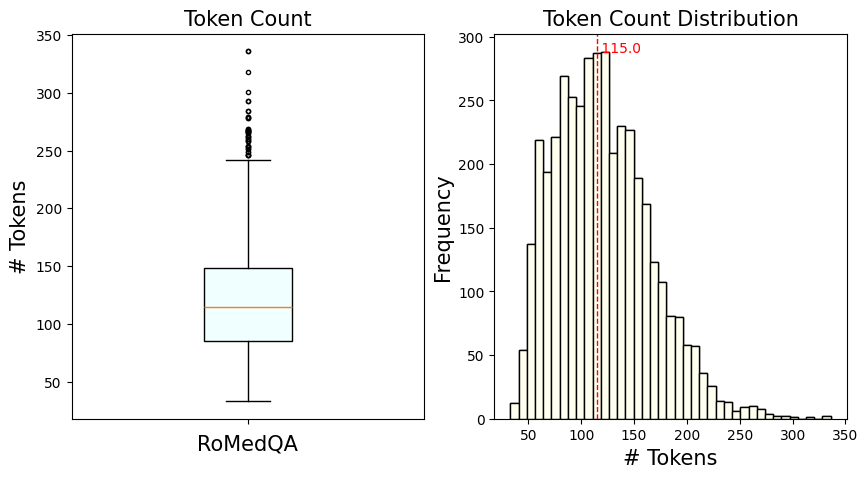}
    \caption{An overview of the sample length distribution regarding the number of tokens.}
    \label{fig:dist}
\end{figure*}

\paragraph{Data Analysis.} Each question has five possible answer choices, numbered from 1 to 5, with only one correct answer. We can notice that the classes are pretty balanced, as depicted in Figure \ref{fig:class}, ensuring that the dataset can be used for potential training and relevant results can be achieved through testing. We first remove stop-words and lemmatize the given entries to compute the TF-IDF scores \cite{sparck1972statistical} in Table \ref{table:tfidf}. We compute the TF-IDF score by multiplying the absolute frequency of each word in the corpus by the logarithm of the IDF. This gives us insights into the dataset's most common subtopics of biology. In Figure \ref{fig:dist}, we show the token length distribution of the dataset. The tokens were computed using the Llama tokenizer.

\section{RoQLlama Training Data Processing}
\label{app:train_ds_processsing}
The dataset used for training RoQLlama contained a significant amount of noise, with samples in Slavic languages being incorrectly labeled as Romanian. Given the large amount of data compared to our limited computing resources, we choose a greedy approach to clean it. We split each text into sentences using the NLTK sentence tokenizer\footnote{https://www.nltk.org/api/nltk.tokenize.html} and then removed any sentences that included characters not based on Latin characters. This step helped eliminate most of the mislabeled text. After that, we combined the cleaned sentences into text samples with fewer than 1,024 tokens for training the model.

\section{Evaluation Prompts}
\label{app:eval_prompts}

In this section, we include the prompts used for testing the Llama2 models, as well as our newly introduced RoQLlama model. This section covers the translated prompts and the prompts used in the Romanian language. These are the prompts used for zero-shot inference. For the few-shot setting, multiple examples are given, one below another, along with the answer keys and format, in the same way the model should respond in the zero-shot scenario.

\subsection{Translated Prompts in English}

\begin{flushleft}
\begin{minipage}{0em}
\fbox{\begin{minipage}{15em}
RoMedQA
\end{minipage}}
\fbox{\begin{minipage}{15em}
I answer medical multiple-choice questions using only the digit of the correct answer from the answer choices. There is only one correct answer. \\

Question: \{\}

Answer:
\end{minipage}}
\end{minipage}
\end{flushleft}

\begin{flushleft}
\begin{minipage}{0em}
\fbox{\begin{minipage}{15em}
RoQA
\end{minipage}}

\fbox{\begin{minipage}{15em}
I read the given context and briefly answer the given questions, using only information from the context. I do not offer any explanation. \\

Context: \{\}

Question: \{\}

Answer:
\end{minipage}}
\end{minipage}
\end{flushleft}

\begin{flushleft}
\begin{minipage}{0em}
\fbox{\begin{minipage}{15em}
REDv2
\end{minipage}}
\fbox{\begin{minipage}{15em}
I read the following text and annotate it based on its predominant emotion.\\

The only emotion categories I can choose from are: Sadness, Surprise, Fear, Anger, Neutral, Trust, and Joy. I do not offer any explanation. \\

Text: \{\}

Emotion:
\end{minipage}}
\end{minipage}
\end{flushleft}

\begin{flushleft}
\begin{minipage}{0em}
\fbox{\begin{minipage}{15em}
RoSTS
\end{minipage}}
\fbox{\begin{minipage}{15em}
I read both sentences and semantically annotate their similarity with a score,  scoring them from 0 (the sentences have no semantic similarity) to 1 (the sentences are identical semantically). I do not offer any explanation. \\

Sentence1: \{\}

Sentence2: \{\}

Semantic similarity score:
\end{minipage}}
\end{minipage}
\end{flushleft}

\begin{flushleft}
\begin{minipage}{0em}
\fbox{\begin{minipage}{15em}
RoMD
\end{minipage}}
\fbox{\begin{minipage}{15em}
I read the following paragraph and annotate it based on its dialect, Romanian or Moldavian.\\

The only dialect categories I can choose from are Romanian and Moldavian. I do not offer any explanation.\\

Paragraph: \{\}

Dialect:
\end{minipage}}
\end{minipage}
\end{flushleft}

\begin{flushleft}
\begin{minipage}{0em}

\fbox{\begin{minipage}{15em}
RoSum
\end{minipage}}
\fbox{\begin{minipage}{15em}
I read the following paragraph and summarize it. \\

Title: \{\}

Paragraph: \{\}

Summary:
\end{minipage}}
\end{minipage}
\end{flushleft}

\begin{flushleft}
\begin{minipage}{0em}
\fbox{\begin{minipage}{15em}
SaRoCo
\end{minipage}}
\fbox{\begin{minipage}{15em}
I read and annotate the following title and paragraph based on the satire category.\\

The only satire categories I can choose from are: satiric and non-satiric. I do not offer any explanation. \\

Title: \{\}

Paragraph: \{\}

Category:
\end{minipage}}
\end{minipage}
\end{flushleft}

\subsection{Original Prompts in Romanian}

\begin{flushleft}
\begin{minipage}{0em}
\fbox{\begin{minipage}{15em}
RoMedQA
\end{minipage}}
\fbox{\begin{minipage}{15em}
Eu răspund la întrebări de medicină de tip grilă doar cu cifra răspunsului corect din variantele de răspuns. Există un singur răspuns corect.\\

Întrebare: \{\}

Răspuns:
\end{minipage}}
\end{minipage}
\end{flushleft}

\begin{flushleft}
\begin{minipage}{0em}
\fbox{\begin{minipage}{15em}
RoQA
\end{minipage}}
\fbox{\begin{minipage}{15em}
Eu citesc contextul dat și răspund succint la întrebările adresate, utilizând doar informații din context. Nu ofer nicio explicație.\\

Context: \{\}

Întrebare: \{\}

Răspuns:
\end{minipage}}
\end{minipage}
\end{flushleft}
\begin{flushleft}
\begin{minipage}{0em}
\fbox{\begin{minipage}{15em}
REDv2
\end{minipage}}
\fbox{\begin{minipage}{15em}
Eu citesc următorul text și îl adnotez în funcție de emoția lui predominanta. \\

Singurele categorii de emoții din care pot sa aleg sunt: Tristețe, Surpriză, Frică, Furie, Neutru, Încredere, Bucurie. Nu ofer nicio explicație. \\

Text: \{\}

Emoție:
\end{minipage}}
\end{minipage}
\end{flushleft}

\begin{flushleft}
\begin{minipage}{0em}
\fbox{\begin{minipage}{15em}
RoSTS
\end{minipage}}
\fbox{\begin{minipage}{15em}
Eu citesc ambele propoziții și adnotez similaritatea semantică dintre cele două propoziții cu un scor de la 0 (propozițiile nu au nicio similaritatea semantică) la 1 (propozițiile sunt identice din punct de vedere semantic). Nu ofer nicio explicație. \\

Propoziție1: \{\}

Propoziție2: \{\}

Scor similaritate semantică:
\end{minipage}}
\end{minipage} 
\end{flushleft}

\vspace{0.7cm}

\begin{flushleft}
\begin{minipage}{0em}
\fbox{\begin{minipage}{15em}
RoMD
\end{minipage}}
\fbox{\begin{minipage}{15em}
Eu citesc următorul paragraf și îl adnotez în funcție de dialectul lui, românesc sau moldovenesc. \\

Singurele categorii de dialect din care pot sa aleg sunt românesc sau moldovenesc. Nu ofer nicio explicație.\\

Paragraf: \{\}

Dialect:
\end{minipage}}
\end{minipage}
\end{flushleft}

\vspace{0.7cm}

\begin{flushleft}
\begin{minipage}{0em}
\fbox{\begin{minipage}{15em}
RoSum
\end{minipage}}
\fbox{\begin{minipage}{15em}
Eu citesc următorul paragraf și îl sumarizez.\\

Titlu: \{\}

Paragraf: \{\}

Sumarizare:
\end{minipage}}
\end{minipage}
\end{flushleft}

\vspace{0.7cm}

\begin{flushleft}
\begin{minipage}{0em}
\fbox{\begin{minipage}{15em}
SaRoCo
\end{minipage}}
\fbox{\begin{minipage}{15em}
Eu citesc următorul titlu și paragraf, și le adnotez în funcție de categoria de satira.\\

Singurele categorii de satiră din care pot sa aleg sunt: satiric sau non-satiric. Nu ofer nicio explicație.\\

Titlu: \{\}

Paragraf: \{\}

Categorie:
\end{minipage}}
\end{minipage}
\end{flushleft}

\end{document}